\documentclass{article}
\usepackage{spconf,amsmath,graphicx}

\usepackage{algorithm,amsfonts}
\usepackage{float}
\usepackage{algpseudocode}
\usepackage[usenames,dvipsnames]{color} 
\usepackage{tabularx} 
\input{latex_resources/mysymbol.sty}
\input{latex_resources/juan_defs.sty}
\usepackage{multirow}
\def\dual{D}

\title{Training Stable Graph Neural Networks\\
Through Constrained Learning}
\name{Juan Cervi\~no, Luana Ruiz and Alejandro Ribeiro \thanks{Support by NSF CCF 1717120, and Theorinet Simons.}}
\address{Department of Electrical and Systems Engineering, University of Pennsylvania, Philadelphia, USA}
\begin{document}
%\ninept
%
\maketitle
\begin{abstract}
Graph Neural Networks (GNN) rely on graph convolutions to learn features from network data. GNNs are stable to different types of perturbations of the underlying graph, a property that they inherit from graph filters. In this paper we leverage the stability property of GNNs as a typing point in order to seek for representations that are stable within a distribution. We propose a novel constrained learning approach by imposing a constraint on the stability condition of the GNN within a perturbation of choice. We showcase our framework in real world data, corroborating that we are able to obtain more stable representations while not compromising the overall accuracy of the predictor.

\end{abstract}
\section{Introduction}
\label{sec:intro}

Graph Neural Networks (GNNs) are deep convolutional architectures tailored to graph machine learning problems \cite{ZHOU202057,gama2018convolutional} which have achieved great success in fields such as biology \cite{NIPS2017_f5077839,NIPS2015_f9be311e} and robotics \cite{qi2018learning,li2019graph}, to name a few. Consisting of layers that stack graph convolutions and pointwise nonlinearities, their successful empirical results can be explained by theoretical properties they inherit from graph convolutions. Indeed, convolutions are the reason why GNNs are invariant to node relabelings \cite{chen2019equivalence,keriven2019universal}; stable to deterministic \cite{gama2018convolutional}, stochastic \cite{gao2021stability}, and space-time graph perturbations \cite{hadou2021space}; and transferable from small to large graphs \cite{ruiz2020graphonTransferability}.

Stability is an especially important property because, in practice, networks are prone to perturbations. For instance, in a social network, friendship links can not only be added or removed, but also strengthened or weakened depending on the frequency of interaction. Similarly, in a wireless network the channel states are dramatically affected by environment noise. Because GNNs have been proved to be stable to such perturbations, in theory any GNN should do well in these scenarios. In practice, however, actual stability guarantees depend on factors such as the type of graph perturbation, the smoothness of the convolutions, the depth and width of the neural network, and the size of the graph \cite{gama2020stability}. In other words, GNNs are provably stable to graph perturbations, but we cannot always ensure that they will meet a certain stability requirement or \textit{constraint}. In improve stability, other works have focused on making GNNs more robust. Some works enforce lipschitz constraints \cite{arghal2021robust}, other arts focus on the architecture of the GNN to make it more robust \cite{jin2020graph}, and others focus on removing noise \cite{luo2021learning}. 

In this paper, our goal is thus to enforce GNNs to meet a specific stability requirement, which we do by changing the way in which the GNN is learned. Specifically, we modify the statistical learning problem by introducing GNN stability as a constraint, therefore giving rise to a constrained statistical learning problem. This leads to an non-convex constrained problem for which even a feasible solution may be may challenging to obtain in practice. To overcome this limitation, we resort to the dual domain, in which the problem becomes a weighted unconstrained minimization problem that we can solve using standard gradient descent techniques. By evaluating the constraint slackness, we iteratively update the weights of this problem. This procedure is detailed in Algorithm \ref{alg:GNNRobust}. In Theorem \ref{thm:NZDG}, we quantify the duality gap, i.e., the mismatch between solving the primal and the dual problems. In Theorem \ref{thm:Convergence}, we present convergence guarantees for Algorithm \ref{alg:GNNRobust}. These results are illustrated numerically in Section \ref{sec:Experiments}, where we observe that GNNs trained using Algorithm \ref{alg:GNNRobust} successfully meet stability requirements for a variety of perturbation magnitudes and GNN architectures.

\section{Graph Neural Networks}
\label{sec:GNN}
A graph is a triplet $\bbG=(\ccalV,\ccalE,\ccalW)$, where $\ccalV=\{1,\dots,N \}$ is its set of nodes, $\ccalE \subseteq \ccalV \times \ccalV$ is its set of edges, and $\ccalW$ is a function assigning weights $\ccalW(i,j)$ to edges $(i,j) \in \ccalE$. A graph may also be represented by the graph shift operator (GSO) $\bbS\in\reals^{N\times N}$, a matrix which satisfies $S_{ij}\neq 0$ if and only if $(j,i)\in \ccalE$ or $i=j$. The most common examples of GSOs are the graph adjacency matrix $\bbA$, $[\bbA]_{ij} = \ccalW(j,i)$; and the graph Laplacian $\bbL = \mbox{diag}(\bbA\boldsymbol{1})-\bbA$. 

We consider the graph $\bbG$ to be the support of data $\bbx=[x_1,\dots,x_N]^\top$ which we call \textit{graph signals}. The $i$th component of a graph signal $\bbx$, $x_i$, corresponds to the value of the data at node $i$. The operation $\bbS\bbx$ defines a \textit{graph shift} of the signal $\bbx$. Leveraging this notion of shift, we define graph convolutional filters as linear shift-invariant graph filters. Explicitly, a graph convolutional filter with coefficients $\bbh=[h_1,\dots, h_{K-1}]^\top$ is given by
\begin{align} \label{eqn:graph_convolution}
    \bby=\bbh *_{\bbS} \bbx = \sum_{k=0}^{K-1}h_k \bbS^k\bbx 
\end{align}
where $*_{\bbS}$ is the convolution operation parametrized by $\bbS$.  

GNNs are deep convolutional architectures consisting of $L$ layers, each of which contains a bank of graph convolutional filters like \eqref{eqn:graph_convolution} and a pointwise nonlinearity $\rho$. Layer $l$ produces $F_l$ graph signals $\bbx_l^f$, called \textit{features}. Defining a matrix $\bbX_l$ whose $f$th column corresponds to the $f$th feature of layer $l$ for $1 \leq f \leq F_l$, we can write the $l$th layer of the GNN as
\begin{equation} \label{eqn:gcn_layer}
\bbX_{l} = \rho \left( \sum_{k=0}^{K-1} \bbS^k \bbX_{l-1} \bbH_{lk} \right) \text{.}
\end{equation}
In this expression, $[\bbH_{lk}]_{gf}$ denotes the $k$th coefficient of the graph convolution \eqref{eqn:graph_convolution} mapping feature $g$ to feature $f$ for $1 \leq g \leq F_{l-1}$ and $1 \leq f \leq F_l$.
A more succinct representation of this GNN can be obtained by grouping all learnable parameters $\bbH_{lk}$, $1 \leq l \leq L$, in a tensor $\ccalH=\{\bbH_{lk} \}_{l,k}$. This allows expressing the GNN as the parametric map $\bbX_L = \phi(\bbX_0,\bbS;\ccalH)$. For simplicity, in the following sections we assume that the input and output only have one feature, i.e., $
\bbX_0 = \bbx \in \reals^N$ and $\bbX_L = \bby \in \reals^N$.

\subsection{Statistical Learning on Graphs}

To learn a GNN, we are given pairs $(\bbx,\bby)$ corresponding to an input graph signal $\bbx \in \reals^N$ and a target output graph signal $\bby \in \reals ^N$ sampled from the joint distribution $p(\bbx,\bby)$. Our objective is to find the filter coefficients $\ccalH$ such that $\phi(\bbx,\bbS;\ccalH)$ approximates $\bby$ over the joint probability distribution $p$. To do so, we introduce a nonnegative loss function $\ell:\reals^N\times\reals^N\to\reals_+$ which satisfies $\ell(\phi(\bbx),\bby)=0$ when $\phi(\bbx)=\bby$. The GNN is learned by averaging the loss over the probability distribution as follows, 

% \begin{subequations}
	\begin{align}
% 	\tag{GL}
	\label{eq:UnconstrainedProblem}
	\underset{\ccalH\in\reals^Q}{\min}     &  \underset{\tiny p(\bbx,\bby)} {\mbE}[\ell \left(\bby,\phi(\bbx,\bbS;\ccalH),\bby)\right)] .
	\end{align}
% \end{subequations}

Problem \ref{eq:UnconstrainedProblem} is the Statistical Risk Minimization problem \cite{shalev2014understanding} for the GNN.
\subsection{Stability to Graph Perturbations}
\label{subsec:Stability}

%GNNs exploit the properties of the underlying graph, thus they are permutation equivariant, which means that a relabeling of the input translates into relabeled output \cite{gama2020stability}. In order to compare the outputs of two GNNs we need to account for this property. By defining the permutation matrices $\ccP$,
%\begin{align}
%    \ccalP = \{ \bbP \in \{0,1\} ^{N \times N} : \bbP \bbone=\bbone, \bbP^\top \bbone =1\},
%\end{align}
%we can thus introduce the \textit{ operator norm modulo permutation } as follows, 
%\begin{align}
%    \|\bbS-\hat \bbS \| _{\ccalP}=\min_{\bbP\in\ccalP} \max_{x:\|x\|=1}\|\bbP^\top \bbS x -  \hat \bbS( \bbP^\top x)\|.
%\end{align}
%where $\ccalP$ is the set of permutation matrixes of $N$ nodes. The importance of the operator distance modulo permutation is that it allows us to measure distances while accounting for the permutation of inputs and outputs that can be generated by the same GNN. 

In the real world, it is not uncommon for graphs to be prone to small perturbations such as, e.g., interference noise in wireless networks. Hence, stability to graph perturbations is an important property for GNNs. Explicitly, we define a graph perturbation as a graph that is $\epsilon$ close to the original graph, 
\begin{equation} \label{eqn:graph_perturbation}
\hat\bbS: \ \|\hat \bbS - \bbS \| \leq \epsilon. 
\end{equation}
An example of perturbation is an additive perturbation of the form $\hat \bbS = \bbS +\bbE$, where $\bbE$ is a stochastic perturbation with bounded norm $\|\bbE\|\leq \epsilon$ drawn from a distribution $\Delta$. 

The notion of GNN stability is formalized in Definition \ref{def:stability}. Note that the maximum is taken in order to account for all possible inputs. 

\begin{definition}[GNN stability to graph perturbations] \label{def:stability}
Let $\phi(\bbx,\bbS;\ccalH)$ be a GNN \eqref{eqn:gcn_layer} and let $\hat\bbS$ be a graph perturbation \eqref{eqn:graph_perturbation} such that $\|\hat\bbS-\bbS\| \leq \epsilon$. The GNN $\phi(\bbX,\bbS;\ccalH)$ is $C$-stable if
\begin{align}\label{eqn:graph_stability}
  \max_{\bbx}  \| \phi(\bbx,\bbS;\ccalH) - \phi(\bbx,\hat\bbS;\ccalH)\| \leq C \epsilon
\end{align} 
for some finite constant $C$.
\end{definition}

A GNN is thus stable to a graph perturbation $\|\hat\bbS-\bbS\|\leq \epsilon$ if its output varies at most by $C \epsilon$. The smaller the value of $C$, the more stable the GNN is to perturbations $\hat \bbS$.
Under mild smoothness assumptions on the graph convolutions, it is possible to show that any GNN can be made stable in the sense of Definition \ref{def:stability} \cite{gama2020stability}. However, for an arbitrary GNN the constant $C$ is not guaranteed to be small and, in fact, existing stability analyses show that it can vary with the GNN depth and width (i.e. number of layers $L$, and number of features $F$ respectively), the size of the graph $N$, and the misalignment between the eigenspaces of $\bbS$ and $\hat\bbS$. What is more, problem \ref{eq:UnconstrainedProblem} does not impose any conditions on the stability of the GNN, thus solutions $\ccalH^*$ may not have small stability bounds $C$. In this paper, our goal is to enforce stability for a constant $C$ of choice. In the following, we show that, on average over the support of the data, better stability can be achieved by introducing a modification of definition \ref{def:stability} as a constraint of the statistical learning problem for the GNN.

\section{Constrained Learning}
\label{sec:CL}

In order to address the stability of the GNN, we can explicitly enforce our learning procedure to account for differences between the unperturbed and the perturbed performance, to do this we resort to the constrained learning theory \cite{chamon2020probably}. We modify the standard unconstrained statistical risk minimization problem (cf. \eqref{eq:UnconstrainedProblem}) by introducing a constraint that requires the solution $\ccalH$ to attain at most an $C \epsilon$ difference between the perturbed and unperturbed problem.

% \begin{subequations}
	\begin{align}
% 	\tag{PGR}
	\label{eq:GraphRobustProblem}
	P^*=\underset{\ccalH\in \reals^Q}{\min}     &  \underset{\tiny p(\bbx,\bby)} {\mbE}[\ell \left(y,\phi(\bbx,\bbS;\bby)\right)] \\
	\text{s.t.}
  &  \underset{\tiny  p(\bbx,\bby,\Delta)}{\mbE} [\ell(\bby,\phi(\bbx,\hat\bbS;\ccalH))-\ell (y,\phi(\bbx,\bbS;\ccalH))]\leq C \epsilon \nonumber
	\end{align}
% \end{subequations}
Note that if the constant $C$ is set at a sufficiently large value, the constraint renders inactive, making problems \eqref{eq:GraphRobustProblem} and \eqref{eq:UnconstrainedProblem} equivalent. As opposed to other methods based on heuristics, or tailored solutions, our novel formulation admits a simple interpretation from an optimization perspective. 
% What is more, in order to solve problem \ref{eq:GraphRobustProblem}, we do not need to model the perturbation, the only requirement is that we are able to sample graphs $\hat\bbS$ according the the distribution $p(\bbx,\bby,\Delta)$.

\subsection{Dual Domain}
In order to solve problem \eqref{eq:GraphRobustProblem}, we will resort to the dual domain. To do so, we introduce the dual variable $\lambda>0 \in \reals$, and we define the Lagrangian function as follows,  
\begin{align}
    \ccalL(\ccalH,\lambda)=&(1-\lambda)\mbE [\ell (\bby,\phi(\bbx,\bbS;\ccalH)]\\ &+\lambda\mbE [\ell (\bby,\phi(\bbx,\hat\bbS;\ccalH))-  \epsilon C].  \nonumber
\end{align}
We can introduce the dual function as the minimum of the Lagrangian $\ccalL (\ccalH,\lambda)$, over $\ccalH$ for a fixed value of dual variables $\lambda$ \cite{boyd2009convex},
\begin{align}
    d(\lambda)=\min_{\ccalH\in \reals^Q} \ccalL(\ccalH,\lambda).
\end{align}
Note that to obtain the value of the dual function $d(\lambda)$ we need to solve an unconstrained optimization problem weighted by $\lambda$. Given that the dual function is a point wise minimum of a family of affine functions, it is concave, even when the problem \eqref{eq:GraphRobustProblem} is not convex. The maximum of the dual function $d(\lambda)$ over $\lambda$ is called the dual problem $D^*$, and it is always a lower bound of problem $P^*$ as follows, 
\begin{align}
    d(\lambda)\leq\dual^*\leq \min_{\ccalH\in\reals^Q} \max_{\lambda\in\reals^+} \ccalL(\phi,\lambda) = P^*.
\end{align}

The advantage of delving into the dual domain, and maximizing the dual function $d$ is that it allows us to search for solutions of problem \ref{eq:GraphRobustProblem} by minimizing an unconstrained problem. The difference between the dual problem $\dual^*$ and the primal problem $P^*$ (cf. \ref{eq:GraphRobustProblem}), is called \textit{duality gap} and will be quantified in the following theorem.

\begin{assumption}\label{as:Lipschitz}
The loss function $\ell$ is $L$-Lipschitz,  i.e.$\|\ell(x,\cdot)-\ell(z,\cdot)\|\leq L\|x-z\|$, strongly convex and bounded by $B$.
\end{assumption}
\begin{assumption}\label{as:NonAtomic} The conditional distribution $p(\bbx,\Delta|\bby)$ is non-atomic for all $\bby\in \reals^N$, and there a finite number of target graph signals $\bby$.
\end{assumption}
\begin{assumption}\label{as:ParameterizationError} There exists a convex hypothesis class $\hat \ccalC$ such that $\ccalC\subseteq\hat \ccalC$, and there exists a constant $\xi>0$ such that $\forall \hat \phi \in\hat \ccalC$, there exists $\ccalH \in \reals^Q$ such that $\sup_{x\in \ccalX}\|\hat\phi(x)-\phi(x,\ccalH) \|\leq \xi$.
\end{assumption}
Note that assumption \ref{as:Lipschitz} is satisfied in practice by most loss functions (i.e. square loss, $L_1$ loss), by imposing a bound. Assumption \ref{as:NonAtomic} can be satisfied in practice by data augmentation \cite{goodfellow2016deep}. Assumption \ref{as:ParameterizationError} is related to the richness of the function class of GNNs $\ccalC$, the parameter $\xi$ can be decrease by increasing the capacity of the GNNs in consideration. To obtain a convex hypothesis class $\hat \ccalH$, it suffices to take the convex hull over the function class of GNNs.

\begin{theorem}[Near-Zero Duality Gap]\label{thm:NZDG}
Under assumptions \ref{as:Lipschitz}, \ref{as:NonAtomic}, and \ref{as:ParameterizationError}, if the Constrained Graph Stability problem  \eqref{eq:GraphRobustProblem} over $\hat \ccalC$ is feasible with constraint slack $\epsilon-\xi$, then the Constrained Graph Stability problem \eqref{eq:GraphRobustProblem}, has near zero duality gap, 

\begin{align}
    P^*-D^*\leq ( \lambda^*+1)L\xi
\end{align}
where $\lambda^*$ is the optimal dual variable of the non-parametric problem with constraint slack $\epsilon-\xi$.
\end{theorem}
The importance of Theorem \ref{thm:NZDG} is that is allows us to quantify the penalty that we incur by delving into the dual domain. Note that this penalty decreases as we make our parameterization richer, and thus we decrease $\xi$. Also note that the optimal dual variable $\lambda^*$ accounts for the difficulty of finding a feasible solution, thus we should expect this value to be small given the theoretical guarantees on GNN stability \cite{gama2020stability}.
\begin{algorithm}[t]
	\caption{Graph Stability Algorithm}
	\label{alg:GNNRobust}
	\begin{algorithmic}[1]
	\State Initialize model $\ccalH^0$, and dual variables $\lambda = 0$
    \For {epochs $e=1,2,\dots$}
    \For {batch $i$ in epoch $e$}
    \State Obtain $N$ samples $\{(\bbx_i,\bby_i)\}_i \sim p(\bbx,\bby)$
    \State Obtain $M$ perturbations $\{(\hat\bbS_i)\}_i\sim \Delta$
    \State Get primal gradient $\nabla_\ccalH \ccalL(\ccalH,\lambda)$ (cf. eq \eqref{eqn:GradientPrimal})
    \State Update params. $\ccalH^{k+1}=\ccalH^{k}-\eta_P \hat\nabla_\ccalH \ccalL (\ccalH^{k},\lambda)$  
    \EndFor
    \State Obtain $N$ samples $\{(\bbx_i,\bby_i)\}_i \sim p(\bbx,\bby)$
    \State Obtain $M$ perturbations $\{(\hat\bbS_i)\}_i\sim \Delta$
    \State Update dual variable $\lambda \leftarrow [\lambda + \eta_D \nabla_\lambda \ccalL(\ccalH,\lambda)]_{+}$
    \EndFor
	\end{algorithmic}
\end{algorithm}

\subsection{Algorithm Construction}
\label{sec:AlgConstruction}

\begin{table*}[]
\centering
\begin{tabular}{c|c|c|c|c|}
\cline{2-5}
                                           & \multicolumn{2}{c|}{RMSE for $1$ Layer GNN}      & \multicolumn{2}{c|}{RMSE for $2$ Layer GNN}      \\ \hline
\multicolumn{1}{|c|}{Norm Of Perturbation} & Unconstrained           & Constrained (Ours)      & Unconstrained          & Constrained (Ours)      \\ \hline
\multicolumn{1}{|c|}{$0$}                  & $0.8631(\pm0.1056)$ & $\mathbf{0.8696(\pm0.0940)}$ & $0.8582(\pm0.1126)$  & $\mathbf{0.8480 (\pm0.1307)}$  \\ \hline
\multicolumn{1}{|c|}{$0.0001$}             & $0.8631(\pm0.1056)$  & $\mathbf{0.8420(\pm0.0884)}$ & $0.8582(\pm0.1126)$ & $\mathbf{0.8273(\pm0.1121)}$ \\ \hline
\multicolumn{1}{|c|}{$0.001$}              & $0.8631(\pm0.1055)$  & $\mathbf{0.8420(\pm0.0884)}$ & $0.8586(\pm0.1128)$ & $\mathbf{0.8273(\pm0.1122)}$ \\ \hline
\multicolumn{1}{|c|}{$0.01$}               & $0.8625(\pm0.1052)$ & $\mathbf{0.8418(\pm0.0877)}$ & $0.8622(\pm0.1147)$  & $\mathbf{0.8277(\pm0.1132)}$ \\ \hline
\multicolumn{1}{|c|}{$0.1$}                & $0.8631(\pm0.1025)$  & $\mathbf{0.8473(\pm0.0807)}$ & $0.9395(\pm0.1374)$ & $\mathbf{0.8320(\pm0.1226)}$  \\ \hline
\multicolumn{1}{|c|}{$0.2$}                & $0.8735(\pm0.0960)$ & $\mathbf{0.8598(\pm0.0831)}$ & $1.1228(\pm0.2291)$ & $\mathbf{0.8483(\pm0.1342)}$ \\ \hline
\multicolumn{1}{|c|}{$0.5$}                & $1.0182(\pm0.1587)$ & $\mathbf{0.9386(\pm0.1079)}$ & $2.0327(\pm0.5074)$   & $\mathbf{0.9500(\pm0.1572)}$ \\ \hline
\end{tabular}
\caption{Evaluations of the RMSE and variance of the trained GNN for $20$ epochs on the testing set (unseen data) for different magnitudes of \textit{relative} perturbations. We consider GNNs of $1$, and $2$ layers with $K=5$ filter taps and $F_1=64$, and $F_2=32$ features for the first and second layer respectively. The constrained learning approach is able to keep a comparable performance on the unperturbed evaluation (i.e., Norm of Perturbation$=0$) while it is more stable as the norm of the perturbation increases.    }\label{table:RMSE_Perturbation}
\end{table*}

In order to solve problem \ref{eq:GraphRobustProblem}, we will resort to iterativelly solving the dual function $d(\lambda)$, evaluate the constraint slack and update the dual variable $\lambda$ accordingly. We assume that the distributions are unknown, but we have access to samples of both graph signals $(\bbx_i,\bby_i)\sim p(\bbx,\bby)$, and perturbed graphs $\hat \bbS_j\sim\Delta$. In a standard learning procedure, to minimize the Lagrangian $\ccalL(\ccalH,\lambda)$ with respect to a set of variables $\lambda$ we can take stochastic gradients as follows, 
\begin{align}\label{eqn:GradientPrimal}
    \hat\nabla_{\ccalH}\ccalL(\ccalH,\lambda)=& \nabla_{\ccalH}\bigg[\frac{1-\lambda}{N}\sum_{i=1}^N \ell(\phi(\bbx_i,\bbS;\ccalH^{k}),\bby_i)\\
    &+\frac{\lambda}{NM}\sum_{i=1}^N\sum_{j=1}^M \ell(\phi(\bbx_i,\hat\bbS_j;\ccalH^{k}),\bby_i)\bigg] \nonumber
\end{align}
The main difference with a regularized problem is that the dual variables $\lambda$ are also updated. To update the dual variables $\lambda$, we evaluate the constraint violation as follows, 
\begin{align}\label{eqn:SubgradientDual}
    \hat\nabla_{\lambda}\ccalL(\ccalH,\lambda)=& \frac{1}{NM}\sum_{i=1}^N\sum_{j=1}^M \ell(\phi(\bbx_i,\hat\bbS_j;\ccalH^{k}),\bby_i) \nonumber \\
    &-\frac{1}{N}\sum_{i=1}^N \ell(\phi(\bbx_i,\bbS;\ccalH^{k}),\bby_i)-C\epsilon.
\end{align}
The intuition behind the dual step is that the dual variable $\lambda$ will increase while the constraint is not being satisfied, adding weight to the stability condition in the minimization of the Lagrangian. Conversely, if the constraint is being satisfied, we will increase the relative weight of the objective function. This means, that if the constraint is more restrictive the optimal dual variable will be larger.

\begin{theorem}[Convergence]\label{thm:Convergence}
Under assumptions \ref{as:Lipschitz}, \ref{as:NonAtomic}, and \ref{as:ParameterizationError}, if for each dual variable $\lambda^k$, the Lagrangian is minimized up to a precision $\alpha>0$, i.e. $\ccalL(\ccalH_{\lambda^k},\lambda^k) \leq \min_{\ccalH\in \reals^Q} \ccalL(\ccalH,\lambda^k)+\alpha$, then for a fixed tolerance $\beta>0$, the iterates generated by Algorithm \ref{alg:GNNRobust} achieve a neighborhood  of the optimal $P^*$ problem in finite time
\begin{align*}
    P^*+\alpha\geq \ccalL(\ccalH^k,\lambda^k)\geq P^*-(2\lambda^*+1)L\xi-\alpha-\beta-\frac{\eta_D B^2}{2}\nonumber
\end{align*}
\end{theorem}
Theorem \eqref{thm:Convergence} allows us to show convergence of Algorithm \ref{alg:GNNRobust} to a neighborhood of the optimal problem \eqref{eq:GraphRobustProblem}.% that depends on the tolerance $\beta$, the precision $\alpha$, the dual step size $\eta_D$ and the loss bound $B$.

\section{Experiments}
\label{sec:Experiments}

% \usepackage{multirow}
% \usepackage{multirow}

We consider the problem of predicting the rating a movie will be given by a user. We leverage the dataset MovieLens 100k \cite{harper2015movielens} which contains $100,000$ integer ratings between $1$ and $5$, that were collected among $U=943$ users and $M=1682$ movies. Nodes in this case are movie ratings, and we construct one graph signal $\bbx_i$ per user $i$. Each entry $[\bbx_i]_j$ storage the rating that user $i$ gave to movie $j$, and $0$ if there is no rating. To obtain the edges, we calculate the movie similarity graph, obtained by computing the pairwise correlations between the different movies in the training set \cite{huang2018rating}. In order to showcase the stability properties of GNNs, we perturb the graph shift operator according to the \textit{Relative Perturbation Modulo Perturbation} \cite{gama2020stability}[Definition 3] model $\hat \bbS = \bbS + \bbE \bbS + \bbS \bbE$. We consider a uniform distribution of $\bbE$ such that, $\|\bbE\|\leq \epsilon$.

We split the dataset into $90\%$ for training and $10\%$ for testing, considering $10$ independent random splits. For the optimizer we used a $5$ sample batch size, and ADAM \cite{Kingma2015AdamAM} with learning rate $0.005, \beta_1=0.9, \beta_2=0.999$ and no learning decay. We used the smooth $L_1$ loss. For the GNNs, we used ReLU as the non-linearity, and we considered two GNNs: (i) one layer with $F=64$ features, and (ii) two layers with $F_1=64$ and $F_2=32$. In both cases, we used $K=5$ filter taps per filter. For the Algorithm \ref{alg:GNNRobust} we used dual step size $\eta_D=0.05$, stability constant $C=0.25$, and magnitude of perturbation $\epsilon=0.2$. For number of perturbation per primal step we used $M=3$, and to evaluate the constraint slackness we used $20\%$ of the training set. 

Table \ref{table:RMSE_Perturbation} shows the RMSE achieved on the test set when the GNN is trained using Algorithm \ref{alg:GNNRobust} and when trained unconstrained for different magnitudes of perturbations of the graph. The numerical results shown in Table \ref{table:RMSE_Perturbation} express the manifestation of the claims that we put forward. First, using Algorithm \ref{alg:GNNRobust} we are able to attain a comparable performance to the one we would have achieved by training ignoring the perturbation. As seen in the first row, the evaluation of the trained GNNs produces comparable results for both the $1$ and the $2$ layer GNN. Second, with our formulation we are able to obtain more stable representations because when the perturbation magnitude increases, the loss deteriorates at a slower rate. This effect is especially noticeable for the $2$ layer GNN. It is well studied that GNN stability worsens as the number of layers increases\cite{gama2020stability,ruiz2020graph}, however using Algorithm \ref{alg:GNNRobust} we are able to curtail this undesirable effect. 

\section{Conclusion}
In this paper we introduced a constrained learning formulation to improve the stability of GNN. By explicitly introducing a constraint on the stability of the GNN we are able to obtain filter coefficients that are more resilient to perturbations of the graph. The benefit of our novel procedure was benchmarked in a recommendation system problem with real world data.
%For future work, we will improve our theoretical guarantees in order to assure stability, and consider other more demanding simulations such as robot swarms. 
\newpage

\bibliographystyle{IEEEbib}
%\bibliography{strings,refs}
\bibliography{bib}

\appendix
\section{Proof of Theorem 1}

Denoting $P_\ccalC$ the primal problem \ref{eq:GraphRobustProblem} over the non-parametric class of functions $\ccalC$, by \cite[Proposition 2]{chamon2020probably}, we express know that, 
\begin{align}
D^* \leq P^*_\ccalC + (1+L\xi )\lambda^* \label{eqn:dualityGapLuiz}
\end{align}
Now noting that $\ccalP \subseteq \ccalC $, it implies that 
\begin{align}
 P^*_\ccalC \leq P^* \label{eqn:Primals}
\end{align}
Combining \eqref{eqn:dualityGapLuiz} and \eqref{eqn:GradientPrimal} attains the desired result. 
\section{Proof of Theorem 2}
See \cite[Theorem 3]{chamon2020probably}.

\end{document}